\def\year{2021}\relax
\documentclass[letterpaper]{article} 
\usepackage{aaai21}  
\usepackage{times}  
\usepackage{helvet} 
\usepackage{courier}  
\usepackage[hyphens]{url}  
\usepackage{graphicx} 
\usepackage{graphicx}  
\usepackage{times}
\usepackage{epsfig}
\usepackage{graphicx}
\usepackage{amsmath}
\usepackage{amssymb}
\usepackage{xspace}
\usepackage{booktabs}
\usepackage{multirow}
\usepackage{subcaption}
\usepackage{url}
\usepackage{xcolor}
\usepackage[normalem]{ulem} 
\usepackage{verbatim}
\usepackage[font=small]{caption}
\usepackage[ruled,vlined,linesnumbered]{algorithm2e} 
\usepackage{tikz}
\usepackage{wrapfig}
\usepackage{sidecap}
\usepackage{adjustbox}
\usetikzlibrary{decorations.pathreplacing,calc}
\usepackage{color, colortbl}
\definecolor{Gray}{gray}{0.9}
\definecolor{LightCyan}{rgb}{0.88,1,1}

\usepackage{microtype}
\usepackage{enumitem}
\newcommand{\etal}{\textit{et al}.}

\usepackage[switch]{lineno}
\title{Meta-Context Transformers for Domain-Specific Response Generation
}
\author{
        Debanjana Kar \thanks{Work done during internship at IBM Research India.}\\
        Dept. of Computer Science\\
        IIT Kharagpur, India\\
        debanjana.kar@gmail.com\\
    \And
        Suranjana Samanta, Amar Prakash Azad\\
        IBM India Research Lab.\\
        Bangalore, India\\
        suransam@in.ibm.com, amarazad@in.ibm.com\\
}

\begin{document}
 \maketitle
\setlength\titlebox{3.8cm}
\begin{abstract}
Despite the tremendous success of neural dialogue models in recent years, it suffers a lack of relevance, diversity, and some times coherence in generated responses. Lately, transformer-based models, such as GPT-2, have revolutionized the landscape of dialogue generation by capturing the long-range structures through language modeling. Though these models have exhibited excellent language coherence, they often lack relevance and terms when used for domain-specific response generation. In this paper, we present DSRNet (Domain Specific Response Network), a transformer-based model for dialogue response generation by reinforcing domain-specific attributes. In particular, we extract meta attributes from context and infuse them with the context utterances for better attention over domain-specific key terms and relevance. We study  the use of DSRNet in a multi-turn multi-interlocutor environment for domain-specific response generation. In our experiments, we evaluate DSRNet on Ubuntu dialogue datasets, which are mainly composed of various technical domain related dialogues for IT domain issue resolutions and also on CamRest676 dataset, which contains restaurant domain conversations. Trained with maximum likelihood objective, our model shows significant improvement over the state-of-the-art for multi-turn dialogue systems supported by better BLEU and semantic similarity (BertScore) scores. Besides, we also observe that the responses produced by our model carry higher relevance due to the presence of domain-specific key attributes that exhibit better overlap with the attributes of the context. Our analysis shows that the performance improvement is mostly due to the infusion of key terms along with dialogues which result in better attention over domain-relevant terms. Other contributing factors include joint modeling of dialogue context with the domain-specific meta attributes and topics.\end{abstract}



\section{Introduction}

Transformer-based pertained language models, such as BERT \cite{devlin2018bert}, GPT-2 \cite{GPT2pre,radford2018improving}, Transformer-XL \cite{Dai2019transformerxl}, XLNet \cite{Yang2019xlnet}, have revolutionized the landscape of natural language processing lately. These models have achieved state-of-the-art performance on many tasks, such as natural language  understanding (NLU), sentence classification, named entity recognition and question answering. The ability to capture the long-range temporal dependencies in the input sequences is one of the key reason behind the success of these models. Besides language coherency, such attributes are beneficial to dialogue response modeling, especially in multi-turn and multi-interlocutor scenarios. The GPT-2 based models \cite{radford2018improving}, pre-trained on a large dataset, have demonstrated that the generated text is fluent, lexically diverse and rich in content. Such models have the capacity to capture textual data with fine granularity and produce output with a high-resolution that closely emulates real-world text written by human.
\begin{figure}
    \centering
    \includegraphics[scale=0.6]{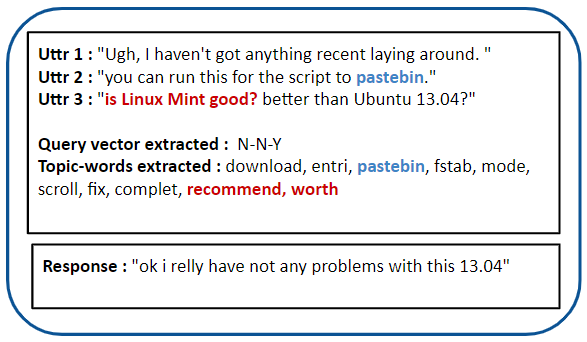}

    \caption{Sample three-turn dialogue instance from the Ubuntu IRC dataset. The highlighted words show the relevance of the topic words with respect to the utterances. "N-N-Y" indicates that the the last utterance is a question, while the previous two are not. This sample highlights several challenges of this dataset : i) The use of domain specific words like 'Ubuntu 13.04', ii) Spelling mistakes like 'relly' in the Response, iii) The query-answer nature of the dialogues, mainly pertaining to the issue resolution channels.}
    \label{fig:my_label}
\end{figure}
Most of the existing neural dialogue response generation models are based on recurrent neural networks(RNNs) 
These neural response generation system suffers from content or style inconsistencies, lack of long-term contextual information \cite{serban17} 
and blandness \cite{li-etal-2016-diversity,Zhang2019ReCoSa}. Many of the issues are alleviated by modeling strategies specifically designed to boost information content using transformer based architecture \cite{GPT2pre} which is evident in some of the recent work on dialogue response generation modeling \cite{Olabiyi2019DLGNet,Bao20soloist,bao20scgpt,zhang20dialogpt}. 

These models have yielded promising results by generating mostly coherent responses given the dialogue context. However, most of them, including the state-of-the-art models trained with naturalistic dialogue data, still perform below human level. Generated responses tend to be either generic, out-of-context or disproportionately short. Some of the previous works attributed such behavior to various causes e.g. prevalence of generic utterances in training data, inadequate sized model architecture to capture the long term temporal dependence, absence of low frequency words in vocab, exposure bias in training models. In a domain specific multi-turn and multi interlocutor dialogue environment, where multiple users converse  over a common channel simultaneously, often regarding a common subject, the above stated problems exacerbate in the generated response. More specifically, the next utterance may not be the response corresponding the immediate previous utterance. In addition, due to multiple agents conversing on the same channel, the context of utterance's might be off to the main content of the discussion. 

In this paper, we propose DSRNet (Domain Specific Response Network), a transformer based model,  where in we alleviate some of the highlighted issues stated above by explicitly inserting meta-context attributes to capture the context better. 
In particular, we extract various meta contexts such as topic, queries etc. (see in Figure \ref{fig:my_label})  and include them as special inputs to our proposed DSRNet. DSRNet architecture is build on GPT-2 \cite{radford2018improving}, with modifications to include the meta contexts in the input set. We have experimented on the Ubuntu dialogue datasets \cite{lowe2015ubuntu,kummerfeld2018large}, which are mainly composed of  various computer domain related dialogues for issue resolutions.  Like GPT-2, DSRNet is formulated as an auto-regressive (AR) language model and uses a multi-layer transformer model architecture. However, unlike GPT-2, DSRNet is trained on large scale dialogues pairs/conversation-sessions extracted from Ubuntu-Dialogue corpus. Our assumption is that this should enable DSRNet to capture the joint distribution over the response and the previous utterance in the conversation flow with finer granularity.  
The input of DSRNet includes the context which constitutes of a predefined number of  previous utterances (before the response) in the conversation, and the meta-contexts. The meta context is composed of conversation topic, query, entities which are extracted from the conversation at hand using traditional NLP approaches. 
We have evaluated DSRNet on the Ubuntu-IRC corpus (multi-interlocutor conversation) \cite{kummerfeld2018large} to generate response utterances which clearly indicate improved response text in terms of alignment with context utterances of the conversation topic. For domain specific environment, it is of great importance to have the response aligned with the context instead of being generic. We have also experimented with other datasets, namely Ubuntu 2.0 dataset (direct conversation) \cite{lowe2015ubuntu} (mainly pertains to the IT  domain) and CamRest676 \cite{wenN2N17} which contains restaurant related conversations. The experimental results on CamRest676 dataset also corroborates improved response text generation. We built DSRNet upon the Huggingface Pytorch Transformer \cite{wolf2019huggingfaces}. We also extended the  Ubuntu-IRC, Ubuntu 2.0  and CamRest676 dataset with meta-contexts. We intend to release both the source code and extended datasets for future research.
%
To the best of our knowledge, our approach is the first to consider explicitly meta-context attributes and leverage it in a transformer based model to generate dialogue responses in a multi-turn dialogue environment. 
The key contributions of our work are as follows:
\begin{itemize}
\item We propose a novel approach, DSRNet, a GPT-2 based model with meta-context attributes for domain specific multi-turn and multi-interlocutor dialogue response generation. 
\item We extend Ubuntu 2.0, Ubuntu-IRC and CamRest676 datasets with meta-context attributes  for better context capturing.
\end{itemize}



\label{intro}

\section{Related Literature}
\label{s:rel}


Pre-trained transformer based language models have shown tremendous advances in the state-of-the-art across a variety of natural language processing (NLP) tasks (\cite{peters18,devlin2018bert,yang19,liu2019roberta,keskar2019ctrl,raffel2019exploring}). These models are often trained to predict words based on their context on massive text data, and the learned models can be fine-tuned to adapt to various downstream tasks. GPT-2 \cite{GPT2pre,radford2018improving} is one of the most known auto-regressive language models and closest to our work which learns language granularity from large amounts of open web text data. Other variants to ground language generation on prescribed control codes are CTRL \cite{keskar2019ctrl} and Grover \cite{zellers19} or latent variables such as Optimus \cite{Li2020OptimusOS}. GPT-2 first investigated massive Transformer-based auto-regressive language models with large-scale text data for pre-training. After fine-tuning, GPT-2 achieves drastic improvements on several generation tasks. One drawback of GPT-2 is the lack of high-level semantic controlling ability in language generation. To alleviate this issue CTRL \cite{keskar2019ctrl} was introduced to train the model based on pre-defined codes such as text style, content description, and task-specific behaviour, mean while Grover \cite{zellers19} generates news article conditioned on authors, dates etc. Unlike these, SC-GPT \cite{bao20scgpt} models the text generation more explicitly which is applied to task-oriented dialogue NLG in a few shot setting. 

In Dialogue domain, several recent works have adapted transformer-based pre-trained language models. DialoGPT \cite{zhang20dialogpt,Wolf2019TransferTransfo} and CGRG \cite{wu2020controllable} extended GPT-2 for chit chat dialogue system. Plato \cite{bao20scgpt} is a pre-trained discrete latent variable model for response generation. For task oriented dialogues, BERT-ToD \cite{wu2020todbert} adapts the pre-trained BERT\cite{devlin2018bert} model to achieve super performance on four dialogue subtasks. SC-GPT \cite{bao20scgpt} and Soloist \cite{Bao20soloist} are  pre-trained models for NLG module that converts a dialogue act into response in natural language. DLGNet \cite{Olabiyi2019DLGNet} is a large transformer model trained on dialogue dataset and achieves good performance on multi-turn dialogue response generation.  Our work is very close to SC-GPT and DLGNet as we also build our model on GPT2 to generate response for a multi-turn dialogues. Moreover, our model reinforces better relevancy by explicit inclusion of context related meta-contexts which helps to capture the context better and its content in the generated responses.

\section{Proposed Approach}
\label{s:approach}

\begin{figure}[!htb]
\begin{center}
\includegraphics[scale=0.35]{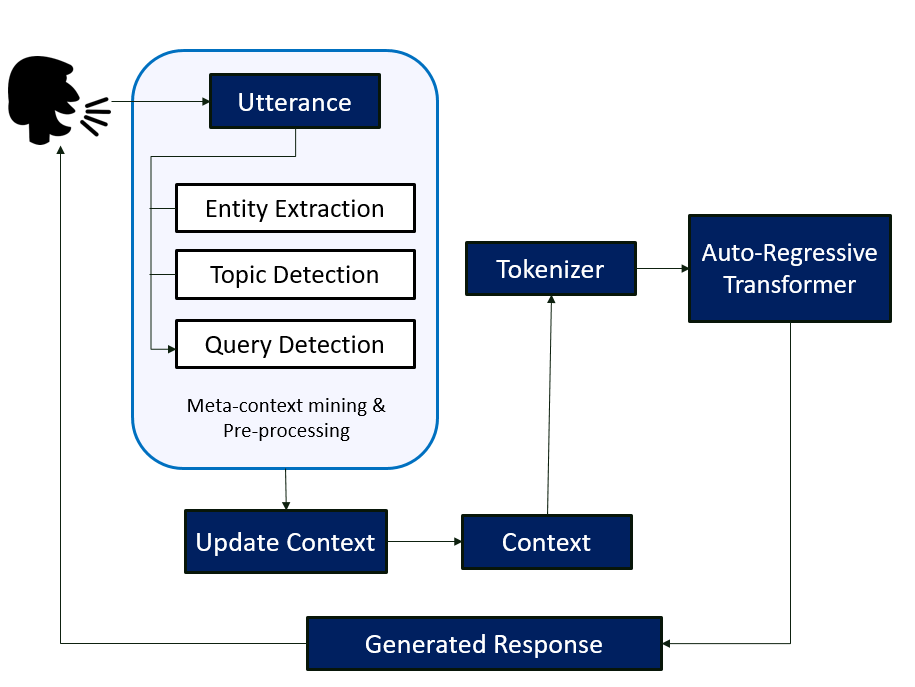} 
\caption{Illustrative overview of the general architecture}
\label{fig.1}
\end{center}
\end{figure}

%
%
We model the task of response generation by incorporating fine-grained meta-contextual attributes to capture domain specific goals in the generated utterances more effectively. We infuse various context attributes, called as meta-context, in our response generation model to capture the content of the conversation context better in the input. Given M training samples $S = {(C_{m}, x_{m})}^{M}_{m=1}$, where $C$ is the input context and $x$ refers to the corresponding response text; our aim is to build a neural model which maximises the likelihood of the generated response conditioned on the context ($C$) and meta-context ($f(C)$), i.e., $p_{\theta}(x|C,f(C))$ parameterized by $\theta$ (model parameters). The decoder being auto-regressive allows us to express the likelihood as
\begin{equation}
p_{\theta}(x|C) = \prod^{W}_{w=1}p_{\theta}(x_{w}|x_{<w}, C, f(C)) 
\label{eq:1}
\end{equation}
where $W$ is the number of tokens in the ground truth response. To capture domain-related terms in a better way, we perform a massive language pre-training, allowing the model to learn domain-related word representations along with their contexts more effectively. The overall flowchart of the proposed method is shown in Figure \ref{fig.1}.
\subsection{Domain language pre-training}
Pre-training on large amounts of data have usually enabled in  better generalization \cite{peng2020few} to newer domains. In our work, we perform pre-training on a massive Ubuntu-dialogue (technical IT support) domain related corpus \footnote{https://github.com/rkadlec/ubuntu-ranking-dataset-creator} with the aim to enrich the model with domain-related knowledge.  To achieve that, we have adopted the pre-trained GPT-2 architecture \cite{radford2018improving} and have trained it further on a language modeling task using the Ubuntu 2.0 Corpora \cite{lowe2015ubuntu}. In this work, we have focused on a corpus which is mainly constituent of technical terms and these words usually aren't part of the general English vocabulary. Through domain language pre-training, we aim to aid the model to learn contextualized representations for such domain-specific words better.
\begin{figure*}[htbp!]
\begin{center}
\includegraphics[scale=0.6]{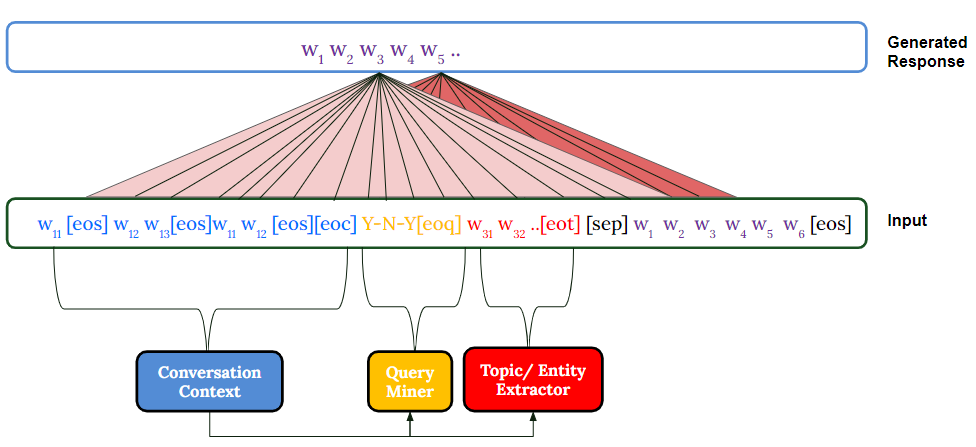} 

\caption{The proposed Meta-Context Transformer. Special tokens [eoc], [eoq], [eot] separate the dialogue history, contextual queries and topics in the context respectively. The token [sep], separates the context from the ground truth that needs to be generated and the token [eos] marks the end of an utterance in our work.}
\label{fig.2}
\end{center}
\end{figure*}

\subsection{Meta-attribute infused fine tuning}
The next utterance in a conversation is usually constructed using key-terms from the context. Most of the prior works on response generation, assumes that the model learns the importance of these key-terms automatically, which may not happen in reality. To tackle this issue of relevance in the generated utterance, we propose infusing meta-attributes, namely topics/entities and queries, from conversations in the training procedure. We process the mined meta-attributes and generate input instances as shown in Table \ref{tab:tr_example}. Thus the meta-context $f(C)$ in Eqn. \ref{eq:1} can be represented as:
\[f(C) = [q_{m};t_{m}]_{m=1}^M\]
where, $M$ is the number of instances in the dataset, $q_m$ represents the query detection result on the $m^{th}$ instance and $t_m$ represents the topic/entity word extracted from the context ($c_m$) of the $m^{th}$ instance. 
In the sections to follow, we provide a detailed description of our meta-attribute mining algorithms. The proposed architecture of DSRNet is illustrated in Figure \ref{fig.2}, where it is highlighted that generation of one word in the response, is dependent explicitly on the context, meta-context and the previous words generated in the response.
\subsubsection{Topic Infusion}
While we are aware of the domain of the corpora, more than often it has been observed that each conversation has a topical premise of it's own. Specially in the Ubuntu related corpora we have used in this work, we have observed conversations involving various topics, mainly pertaining to troubleshooting, version releases - to name a few. In our work, we adopt an unsupervised approach to extract the dominant topic of each utterance. 
In particular, we have used Latent Dirichlet Allocation \cite{blei2003latent} method to determine the topics of the conversations in the corpus. We have treated each conversation in the corpus as a document in our approach. The number of topics rendering the optimal coherence score was found to be 80 for the Ubuntu IRC data-set. Using the learnt topic model, we have used it to determine the dominant topic of each utterance and have appended the 10 most contributing topic keywords from the corpus to each utterance. Using the statistical approach of topic modeling has not only saved us annotation labour, it has also supported the model with consistent performance as reported in Tables \ref{tab:camrest}-\ref{tab:kummerfeld}. 

\subsubsection{Query Mining and Infusion }
Query mining module deals with identifying queries from context utterances. We observe that query utterances, if present, directly influences the conversation responses utterance. We infuse the identified queries as a part of meta-attributes in the form of query vector as shown in Figure\ref{fig:my_label}. We note that the queries are typically the last utterance prior to response while fewer times it occurs at some prior utterances. Queries, or  question utterances, can take both explicit and implicit lexical forms \cite{Mckewon2004,Forsythand2007},
 e.g., explicit: \textit{"What is the latest Ubuntu version?"}, implicit: \textit{"I was wondering what is the latest Ubuntu version released."}. Furthermore, the queries may also contain informal utterance construct as the conversations are informal in nature. 
To identify the queries, we adopted a semi- supervised approach where we augmented some lexical rules along with an SVM model. The lexical rules are apt to capture queries containing question words (mainly constituting 5W1h question words, for e.g.'?', when, how, where), along with a set of other curated verb and adverb based question words (e.g.,could, did, kindly, please). On the other hand, the  SVM model is trained on the NPS Chat dataset \footnote{http://faculty.nps.edu/cmartell/NPSChat.htm} to detect the implicit queries which captures the informal query utterances. Our algorithm identifies an utterance as negative query only when both lexical and SVM model yields negative label.

We have randomly sampled 100 instances from the Ubuntu datasets and have evaluated our query mining algorithm. We report a precision score of 86.90\%, a recall of 92.03\% and an F-Score of 89.39\% . We feed each utterance in the context to the query mining algorithm to determine if it is a query or not. Based on the decisions obtained from the algorithm, we append a sequence of 'Y-N's to the context - 'Y's confirming the presence of a query at that index in the context and 'N's confirming otherwise. Sample training instances depicted in Table \ref{tab:tr_example} provide a clearer picture of the input format.

\subsubsection{Entity Extraction}
Although we capture the topical preference of the conversation by infusing topic keywords with each utterance in the context, we realise that the domain-keywords in the context may still not be captured in the generation. This is mainly because i) the topic keywords do not necessarily pertain to that particular  conversational context, ii) it is often tricky to capture the optimal number of topics for topic modeling and may result in topic overlaps - hence, losing out on important topic words, iii) dialogues are known to be privy to noisy text - with spelling mismatches and word distortions, some important domain words in the context may not be captured accurately. To tackle these issues, we experiment with entity phrases extracted from the context, instead of topic words and report the performance in Tables \ref{tab:camrest}-\ref{tab:kummerfeld}. 
To extract entity words from conversations we adopt the weakly supervised method as outlined in \cite{mohapatra2018domain}.

\section{Dataset}
\label{s:data}
We have used three popular, publicly available datasets to train and evaluate our proposed model DSRNet. While the Ubuntu 2.0 dataset \footnote{https://github.com/rkadlec/ubuntu-ranking-dataset-creator} \cite{lowe2015ubuntu} and the Ubuntu IRC dataset \cite{kummerfeld2018large} mainly pertain to the technical domain, the CamRest676 dataset \cite{wenN2N17} caters to the restaurant enquiry task. For the domain language pre-training step, our dataset of choice was the Ubuntu 2.0 dataset due to it's massive collection of annotated examples. 
It is important to note, that natural language generation for domains like restaurant, hotels and travel are much  simpler than the same in a technical domain. Dialogues in both Ubuntu 2.0 and Ubuntu IRC dataset mainly constitute technical words like Ubuntu terminal commands (eg. sudo apt-get), links or URLs to solutions, technical jargon (e.g. html, css ) which traditionally do not exist in general English vocabulary. Certain words like \textit{Windows, network}, although they do exist in the English vocabulary, they have a very different meaning in a technical domain. We tackle most of these challenges through our domain language pre-training and meta-attribute learning steps. Moreover, the dialogues in the Ubuntu IRC dataset occur in a multi-turn, multi-locutor setting. This means that in a conversation instance there can be multiple parallel sub-conversations taking place. We adhere to the following pre-processing rules to tackle some of the challenges of the dataset : i) We extract the conversations from the dataset using the disentanglement annotations provided (for the Ubuntu IRC dataset), and we remove, ii) bad words from the dialogues - we do not want our system to learn foul language, iii) usernames, timestamps from utterances, iv) non-English utterances from the corpus (we encountered a number of Spanish utterances in the conversations), v) instances of utterance repetitions, and vi) one word utterances which are neither questions nor commands.

\section{Experimental Studies}
\label{s:exp}

\begin{table*}
    \centering
    \begin{tabular}{|p{1in}|p{6in}|}
    \hline
     Dataset & Training example  \\ \hline
     Ubuntu 2.0 (with entities) &  I already do [eos] Ok, goto the /root folder and control+hThat'll show the hidden foldersThat'll show the hidden foldersand the trash folder should be in a .gnome folder [eos] so how do I get permissions to open root? [eos]  [eoc] N-N-Y [eoq] root, permission, trash folder, hide foldersand [eot] [sep] Press Alt+f2 and type this in: gksudo nautilus [eos]  \\ \hline
     
     Ubuntu IRC (with entities) & however, in addition, I've observed something odd with this machine [eos] Sorry, I've had a quick look over a few Ubuntu mirrors, but can't find a Live PPC CD of Warty... are they available? [eos] any reboot short of turning the mains power off causes the bios to fail to recognize the hard drive [eos][eoc] N-Y-N [eoq] available, warty, machine, addition, bios, reboot, live ppc cd, few ubuntu mirror, quick look, something odd, hard drive, main power [eot] [sep] there's no ppc live cd yet [eos] \\ \hline
     Ubuntu IRC (with topics) & however, in addition, I've observed something odd with this machine [eos] Sorry, I've had a quick look over a few Ubuntu mirrors, but can't find a Live PPC CD of Warty... are they available? [eos] any reboot short of turning the mains power off causes the bios to fail to recognize the hard drive [eos][eoc] N-Y-N [eoq] connect, fail, reinstal, final, enter, normal, command, address, account, mode [eot] [sep] there's no ppc live cd yet [eos] \\ \hline
     CamRest676 (with entities) & I need to find an expensive restauant that's in the south section of the city. [eos] There are several restaurants in the south part of town that serve expensive food. Do you have a cuisine preference? [eos] No I don't care about the type of cuisine. [eos]  [eoc] N-Y-N [eoq] city, town, type, south section, expensive restauant, expensive food, south part, several restaurant, cuisine preference [eot] [sep] Chiquito Restaurant Bar is a Mexican restaurant located in the south part of town. [eos] \\ \hline
     CamRest676 (with topics) & I need to find an expensive restauant that's in the south section of the city. [eos] There are several restaurants in the south part of town that serve expensive food. Do you have a cuisine preference? [eos] No I don't care about the type of cuisine. [eos][eoc] N-Y-N [eoq] cheap, moder, south, price, princ, thanh, postcod, gastropub, hous, turkish [eot] [sep] Chiquito Restaurant Bar is a Mexican restaurant located in the south part of town. [eos] \\ \hline
    \end{tabular}
    \caption{Examples of training samples from Ubuntu 2.0, Ubuntu-IRC and CamRest676 dataset used in DSRNet.}
    \label{tab:tr_example} 
\end{table*}

In this section, we evaluate our proposed model DSRNet on three different dialog corpus. We address the following questions during the evaluation process, which are (i) How efficient is DSRNet when compared with other state-of-the-art (SOTA) methods of generating response in dialog settings and (ii) How efficiently DSRNet is able to improve the task of response generation using the meta-contexual information.

\subsection {Experimental settings}
We fine-tune the pre-tuned GPT-2 model on our dataset using 2 v100 cores using 100 GPU memory in each of the them, for most of the experimental settings.
\subsubsection{Dataset for fine-tuning}
We create the training instances by considering contexts in a sliding window fashion, containing three consecutive utterances. The context is followed by the query feature obtained from query mining. This represents which of the utterances in the context is a question (marked by Y) and which is not (marked by N). This helps to give different priority to questions which are present in the context. This is followed by (i) the list of entities extracted from the context or (ii) list of 10 dominant topic words that represent the context. Typical examples of a training instance, using both query information and entities is shown in Table \ref{tab:tr_example}.

Ubuntu2.0 corpus has about 3 million training instances, when the context has 3 utterances. Camrest676 has $2515$ training instances and Ubuntu-IRC dataset has 22582 training instances. We use 10,000 test samples for Ubuntu 2.0 dataset and 1000 test samples from each of the other two dataset. Ubuntu IRC corpus \cite{kummerfeld2018large} has 153 conversation files, each of which have several parallel conversations. We use the annotation provided by Kummerfeld \etal and extract $4621$, $392$ and $298$ conversations for training, evaluation and test split respectively. The entire dataset with meta contextual information will be made available online for research purpose.

\subsubsection{Topic modeling}
\begin{figure}
\begin{center}
\includegraphics[scale=0.25]{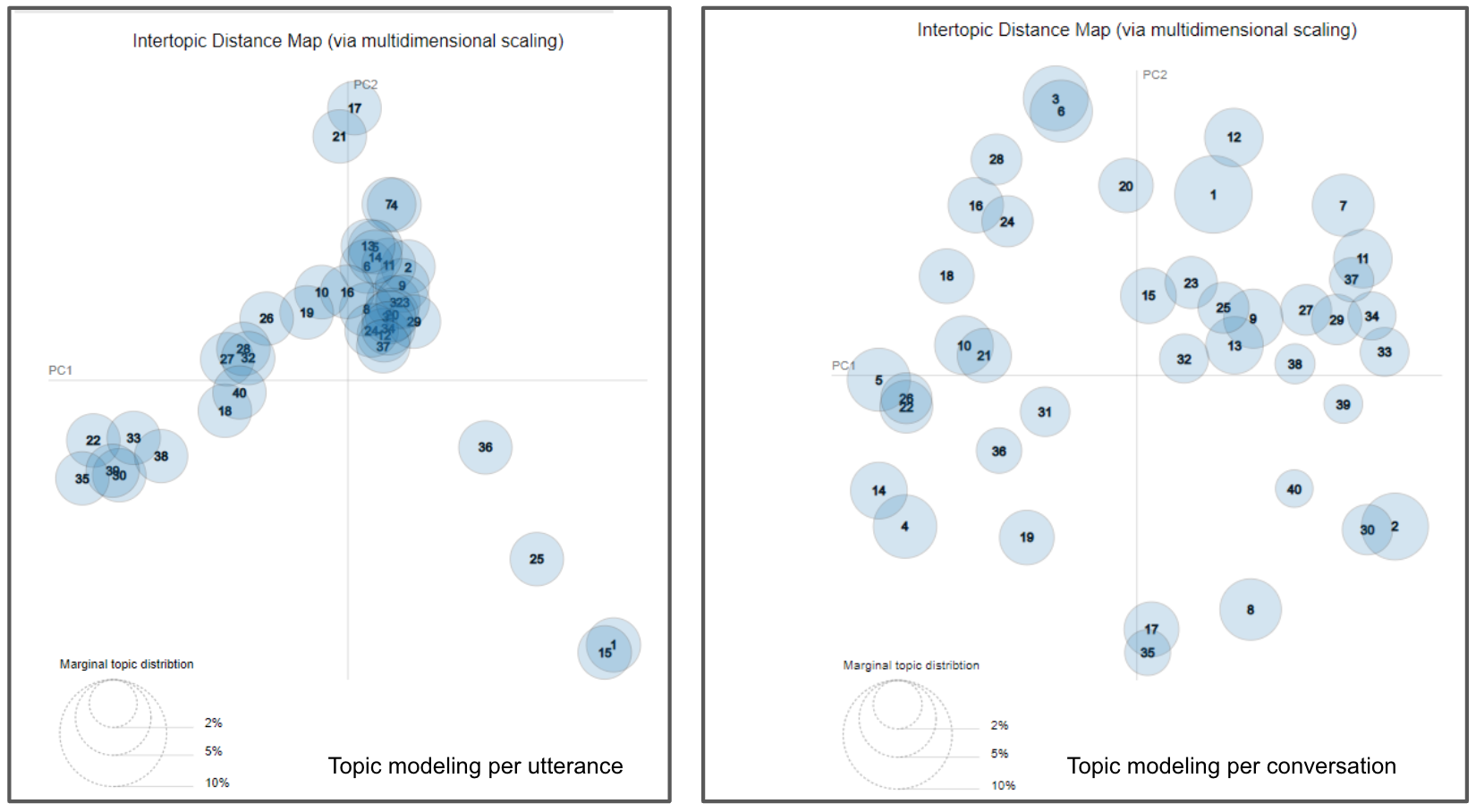} 

\caption{Figure shows the topic distribution when topic modeling is done at (i) utterance level (left) and (ii) conversation level (right) with number of topics set as 40.}
\label{fig:topic_modeling}
\end{center}
\end{figure}

We do stemming of the words in the sentences and consider each conversation as one document. We apply a standard LDA based topic-modeling to extract the dominant topics in the corpus. Figure \ref{fig:topic_modeling} shows the topic distribution (number of topics 40 in this case, for better visual aid) at utterance level and conversation level respectively. It is clear from the figure that better discriminating topics are obtained at the conversation level. We set the number of topics to 80 in all the three corpus, after observing the coherence score.

\subsubsection{Evaluation Metrics}
We use the nlg-eval toolkit \footnote{https://github.com/Maluuba/nlg-eval} and MultiTurnDialogZoo toolkit \footnote{https://github.com/gmftbyGMFTBY/MultiTurnDialogZoo} for obtaining  the values for the following metrics during our experimentation:
\begin{enumerate}
    \item BLEU and ROUGE - BLEU has been used for measuring the coherency of the generated output. It mainly looks at the precision of the common n-grams in the ground truth and the generated output. Similarly, Rouge measures the recall of the common n-grams occurring in ground truth and generated output.
    \item DISTINCT - It measures the diversity of the words in the generated output. We consider the average of DISTINCT-1 and DISTINCT-2 scores.
    \item BERT-Score (BERTSc) - BERT score represents the cosine similarity of the pair-wise similar words in the embedded space.
\end{enumerate}
We would like to mention, that in spite of having so many evaluation metrics, it is difficult to compare the generated response with the ground truth, as an utterance can be represented correctly with many different sentences. Many a times, human evaluation is the best way to determine the correctness of a generated output. 

\subsubsection{Baseline}
The following methods have been adapted as baselines which gives SOTA results on response generation in dialog settings.
\begin{enumerate}
    \item   Vanilla GPT-2 - We use the GPT-2, finetuned on the training set, as one of the baselines. As the proposed DSRNet is an improved version of GPT-2, it gives an idea about how strong the modification of the proposed method is with respect to the base model.
    \item HRED, VHRED - We train \cite{aless2015hierarchical} and \cite{Serban2016BuildingED} on each of the datasets for 20 epochs. Both of the methods model the task using recurrent encoders and decoders in a hierarchical fashion.
    \item DLGNet - We consider DLGNet for comparison purpose for the Ubuntu 2.0 dataset. The evaluation metric values are directly taken from the paper.
\end{enumerate}

\subsection{Performance and evaluation}
\label{l:perf}
We choose the training parameters based on the performance on the evaluation data split and report the metric results on the test split. For the grid search of hyper parameters, we use 6 different randomly generated seed value, 2 sequence lengths of $120$ and $150$ and $2$ learning rates of $5e-5$ and $1e-4$. We show the results of different variations of DSRNet - (i) DSRNet (qstn) - having question detection only as the meta-context (ii) DSRNet (qstn+top) - having question detection and dominant topic words as meta context, and (iii) DSRNet (qstn+ent) - having question detection and entities extracted from context as meta context. We keep topic words and entities as complimentary information in forming the meta context of an instance. The results for three dataset are explained below.

We fine-tune the proposed DSRNet on CamRest676 dataset for $20$ epochs. Table \ref{tab:camrest} shows the performance of different models on generating response for CamRest676 dataset. We compare the performance of DSRNet with that of vanilla GPT-2, HRED and VHRED.  The optimal parameters used for the experiments to obtain metric scores of DSRNet are as follows: learning rate - $5e-5$, sequence length - $120$ and seed-value - $198$.

\begin{table}[!htbp]
    \centering
    \begin{tabular}{|p{1.5cm}|c|c|c|c|}
    \hline
     Method & BLEU & ROUGE & BERTSc &  Distinct  \\ \hline
     HRED & 0.0332 &  0.0582 & -0.0007 &  0.0655 \\ 
     VHRED & 0.0298 &  0.0722 & 0.0049 &  0.0667 \\ 
     GPT-2 & 0.1382  & 0.1496 & 0.242 & 0.4164 \\ 
     DSRNet (qstn) & 0.211  & 0.2581 & 0.3296  & 0.4356 \\
     DSRNet (qstn+top) & 0.1896  & 0.1679 & 0.2254 & 0.4002 \\
     DSRNet (qstn+ent) & \textbf{0.260}  & \textbf{0.2587} & \textbf{0.6603} & \textbf{0.4533} \\ \hline
    \end{tabular}
    \caption{Performance evaluation on CamRest676 dataset (best in bold).}
    \label{tab:camrest}
\end{table}

Table \ref{tab:ubuntu} shows the performance of different models on generating response for Ubuntu2.0 dataset. We compare the performance of DSRNet with that of vanilla GPT-2 and DLGNet. The optimal parameter settings for this experimentation are: learning rate: $5e-05$, batch size: $4$, sequence length: $120$ and seed value: $906$. We can observe that DLGNet is having better BLEU score than DSRNet, which is mostly because of the fact that DLGNet considered a much larger sequence length (1024) while training their model. This is particularly helpful for Ubuntu 2.0 dataset, where the length of utterance can be very long.

\begin{table}[h]
    \centering
    \begin{tabular}{|p{1.5cm}|c|c|c|c|}
    \hline
     Method & BLEU & ROUGE & BERTSc &  Distinct  \\ \hline
     HRED & 0.017&  0.0483& NA &  0.046 \\ 
     VHRED & 0.0171&  0.0855& NA &  0.089 \\ 
     DLGNet & \textbf{0.0279} &  \textbf{0.2191}& - &  0.4953 \\ 
     GPT-2 & 0.0122  & 0.0241 & 0.0138 & 0.2023 \\ 
     DSRNet (qstn) & 0.0153  & 0.0230 & 0.0341  & 0.4797 \\
     DSRNet (qstn+top) & 0.0155  & 0.0237 & 0.0335 & 0.5051 \\
     DSRNet (qstn+ent) & 0.0197  & 0.0321 & \textbf{0.0328} & \textbf{0.5561} \\
     \hline
    \end{tabular}
    \caption{Performance evaluation on Ubuntu2.0 dataset.}
    \label{tab:ubuntu}
\end{table}

Table \ref{tab:kummerfeld} shows the performance of different models on generating response for Ubuntu-IRC dataset. The optimal set of training parameters used in this experiments are: learning rate - $5e-05$, batch size: 4, sequence length: 150 and seed value: 606.

\begin{table}[htbp]
    \centering
    \begin{tabular}{|p{1.5cm}|c|c|c|c|}
    \hline
     Method & BLEU & ROUGE & BERTSc &  Distinct  \\ \hline
      HRED &0.118  &0.11  &0.07  & 0.00025\\ 
      VHRED &0.111  &0.21  &0.21  & 0.0022\\ 
      GPT-2 & 0.0854  & 0.1032 & 0.2354 & 0.2194 \\ 
      DSRNet (qstn) & 0.1062  & 0.1451 & 0.2879  & 0.4676 \\
      DSRNet (qstn+top) & 0.1287  & \textbf{0.1890} & 0.2981 & 0.4432 \\
      DSRNet (qstn+ent) & \textbf{0.1483}  & 0.1566 & \textbf{0.3472} & \textbf{0.4701} \\ \hline
    \end{tabular}
    \caption{Performance evaluation on Ubuntu-IRC dataset.}
    \label{tab:kummerfeld}
\end{table}

It is evident from the tables that adding of meta contextual information does help in generating better responses. Addition of entities is performing better than that of topic words, as entities are directly picked up from the corresponding context, whereas, a dominant topic word may come from outside of the context. It is also to be noted, that the repetition of generated response for multiple similar context is least for DSRNet, which is evident from the DISTINCT score. Also, as per our observation, DSRNet produced more semantically and syntactically meaningful responses.

\subsection{Importance of Query mining and Entity Extraction}
\label{l:importance}
\begin{figure}
\begin{center}
\includegraphics[scale=0.35]{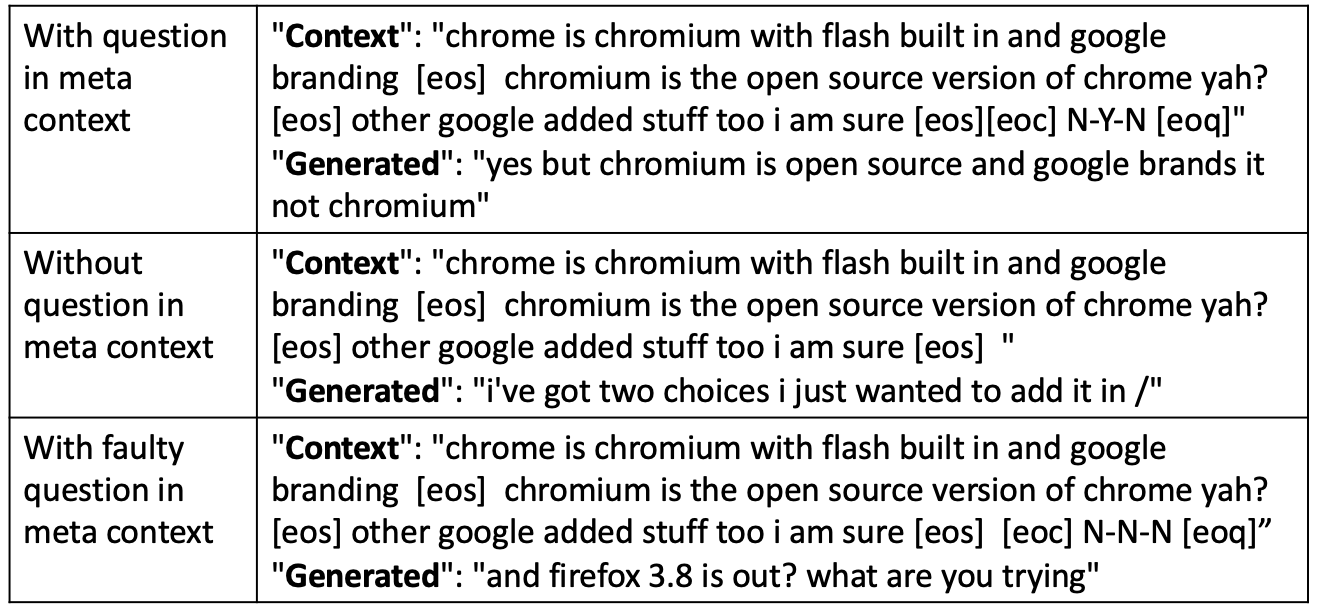}
\caption{Examples of generated response from DSRNet with and without using questions as meta context.}
\label{fig:question}
\end{center}
\end{figure}

\begin{figure}
\begin{center}
\includegraphics[scale=0.35]{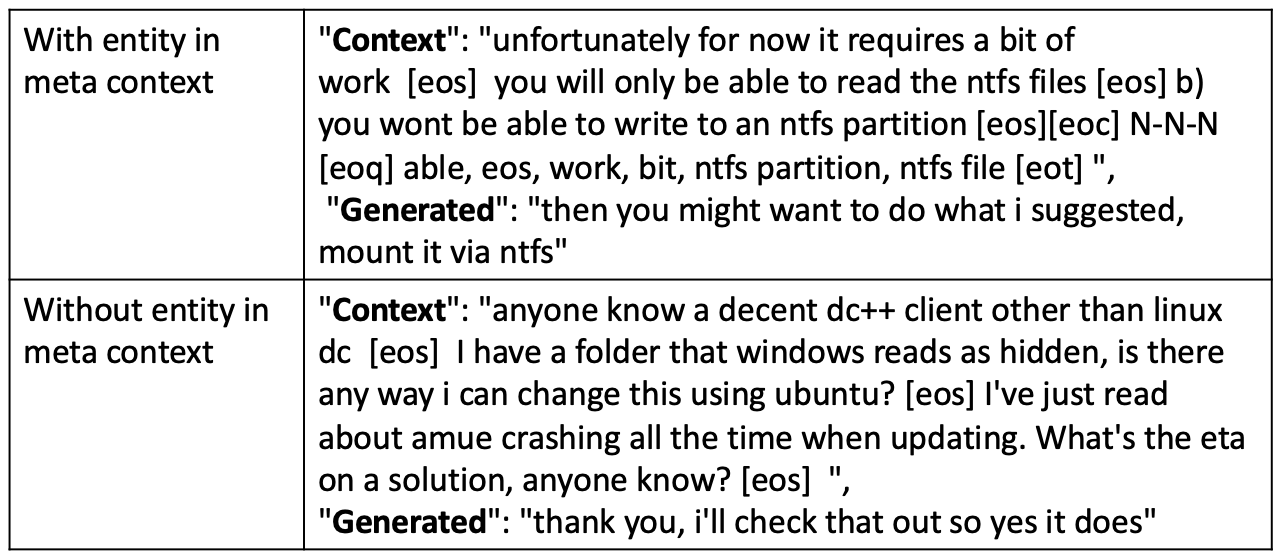}
\caption{Examples of generated response from DSRNet with and without using entities as meta context.}
\label{fig:entity}
\end{center}
\end{figure}

The idea of including the query mining in the meta-context is that the questions in the context should get more importance while generating the response. This is particularly true in support chats, where the questions being asked needs to be answered without much delay. We observed the response generated without using the question mining in the context and also when the question detection module gives erroneous results. In both these cases, the quality of the generated response degrades, as seen in Figure \ref{fig:question}.
Similarly, the evaluation metrics for all three dataset show the improved performance of DSRNet when entities are being used in the meta context. Figure \ref{fig:entity} shows an example, where the generated response contains the entities in it.

\subsection{Ablation study}
Most of the time the generated response of DSRNet is meaningful to the context, with proper grammar. However, in general sentences, for example, in case of closing remarks in a conversation, the generated output suffers from repetition of words and sentences. Many a times, for CamRest676 dataset, while concluding a conversation, the generated output response from DSRNet is given as: \textit{"no, thank you have a nice day! thank you and goodbye. you and goodbye"}. Most of the generation models too suffer from repeatability problem. We believe that this problem is mostly due to the small size of the dataset.

There has been some responses, which end abruptly, such as \textit{"there are two moderately priced mid-range restaurants, both in the centre part of town. do you"}. This is mainly because of the sequence length that we are considering during training of DSRNet, due to computational effectiveness of the model training. With a large value of sequence length, this problem can be resolved.

There are also some of the generated responses, which are highly relevant with the context, but is different from that of the ground truth utterance, such as: 

\noindent \textit{"Context": "There are several good restaurants in the south part of town. Do you have a preference for the type of food or price range? [eos] Yes, I'd like a restaurant that serves portuguese food. [eos] nandos is the only restaurant i can find in the south that serves portuguese. [eos] [eoc] Y-N-N [eoq] town, type, preference, nandos, south part, price range, portuguese food, restaurant [eot]"}
        
\noindent \textit{"Generated": "what is the address?"}
        
\noindent \textit{"True": "Nandos sounds great. Thank you."}

\noindent These generated outputs will reduce the metric scores used for evaluation, but a human evaluation may consider the generated output as a relevant one.

\section{Conclusion}
\label{s:conclusion}


In this paper, we have proposed DSRNet, a transformer-based model, for dialogue response generation by explicitly infusing domain-specific attributes. To infuse meta context in DSRNet, we have extracted meta attributes, namely conversation topics, key entities and queries from the conversation context which enables better relevance of the generated response.
%
We conducted thorough investigation over Ubuntu-IRC, Ubuntu 2.0, and CamRest676 dataset and reported multiple performance metrics, such as BLEU, ROGUE, and semantic similarity scores. 
Our evaluation results indicate that DSRNet shows improvement with other existing models in terms of generating responses which are more relevant to the conversation context.  The generated responses have better presence of domain-specific key attributes that exhibit better overlap with the attributes of the context. 




\bibliographystyle{aaai}
\bibliography{main_v0}

\begin{thebibliography}{}

\bibitem[\protect\citeauthoryear{Blei, Ng, and Jordan}{2003}]{blei2003latent}
Blei, D.~M.; Ng, A.~Y.; and Jordan, M.~I.
\newblock 2003.
\newblock Latent dirichlet allocation.
\newblock {\em Journal of machine Learning research} 3(Jan):993--1022.

\bibitem[\protect\citeauthoryear{Dai \bgroup et al\mbox.\egroup
  }{2019}]{Dai2019transformerxl}
Dai, Z.; Yang, Z.; Yang, Y.; Carbonell, J.; Le, Q.~V.; and Salakhutdinov, R.
\newblock 2019.
\newblock Transformer-xl: Attentive language models beyond a fixed-length
  context.
\newblock In {\em ACL}.

\bibitem[\protect\citeauthoryear{Devlin \bgroup et al\mbox.\egroup
  }{2018}]{devlin2018bert}
Devlin, J.; Chang, M.-W.; Lee, K.; and Toutanova, K.
\newblock 2018.
\newblock Bert: Pre-training of deep bidirectional transformers for language
  understanding.

\bibitem[\protect\citeauthoryear{{Forsythand} and
  {Martell}}{2007}]{Forsythand2007}
{Forsythand}, E.~N., and {Martell}, C.~H.
\newblock 2007.
\newblock Lexical and discourse analysis of online chat dialog.
\newblock In {\em International Conference on Semantic Computing (ICSC 2007)},
  19--26.

\bibitem[\protect\citeauthoryear{Keskar \bgroup et al\mbox.\egroup
  }{2019}]{keskar2019ctrl}
Keskar, N.~S.; McCann, B.; Varshney, L.~R.; Xiong, C.; and Socher, R.
\newblock 2019.
\newblock Ctrl: A conditional transformer language model for controllable
  generation.

\bibitem[\protect\citeauthoryear{Kummerfeld \bgroup et al\mbox.\egroup
  }{2018}]{kummerfeld2018large}
Kummerfeld, J.~K.; Gouravajhala, S.~R.; Peper, J.; Athreya, V.; Gunasekara, C.;
  Ganhotra, J.; Patel, S.~S.; Polymenakos, L.; and Lasecki, W.~S.
\newblock 2018.
\newblock A large-scale corpus for conversation disentanglement.
\newblock {\em arXiv preprint arXiv:1810.11118}.

\bibitem[\protect\citeauthoryear{Li \bgroup et al\mbox.\egroup
  }{2016}]{li-etal-2016-diversity}
Li, J.; Galley, M.; Brockett, C.; Gao, J.; and Dolan, B.
\newblock 2016.
\newblock A diversity-promoting objective function for neural conversation
  models.
\newblock In {\em Proceedings of the 2016 Conference of the North {A}merican
  Chapter of the Association for Computational Linguistics: Human Language
  Technologies},  110--119.
\newblock San Diego, California: Association for Computational Linguistics.

\bibitem[\protect\citeauthoryear{Li \bgroup et al\mbox.\egroup
  }{2020}]{Li2020OptimusOS}
Li, C.; Gao, X.; Li, Y.; Li, X.; Peng, B.; zhe Zhang, Y.; and Gao, J.
\newblock 2020.
\newblock Optimus: Organizing sentences via pre-trained modeling of a latent
  space.
\newblock {\em ArXiv} abs/2004.04092.

\bibitem[\protect\citeauthoryear{Liu \bgroup et al\mbox.\egroup
  }{2019}]{liu2019roberta}
Liu, Y.; Ott, M.; Goyal, N.; Du, J.; Joshi, M.; Chen, D.; Levy, O.; Lewis, M.;
  Zettlemoyer, L.; and Stoyanov, V.
\newblock 2019.
\newblock Roberta: {A} robustly optimized {BERT} pretraining approach.
\newblock {\em CoRR} abs/1907.11692.

\bibitem[\protect\citeauthoryear{Lowe \bgroup et al\mbox.\egroup
  }{2015}]{lowe2015ubuntu}
Lowe, R.; Pow, N.; Serban, I.; and Pineau, J.
\newblock 2015.
\newblock The ubuntu dialogue corpus: A large dataset for research in
  unstructured multi-turn dialogue systems.
\newblock {\em arXiv preprint arXiv:1506.08909}.

\bibitem[\protect\citeauthoryear{Mohapatra \bgroup et al\mbox.\egroup
  }{2018}]{mohapatra2018domain}
Mohapatra, P.; Deng, Y.; Gupta, A.; Dasgupta, G.; Paradkar, A.; Mahindru, R.;
  Rosu, D.; Tao, S.; and Aggarwal, P.
\newblock 2018.
\newblock Domain knowledge driven key term extraction for it services.
\newblock In {\em International Conference on Service-Oriented Computing},
  489--504.
\newblock Springer.

\bibitem[\protect\citeauthoryear{Olabiyi and Mueller}{2019}]{Olabiyi2019DLGNet}
Olabiyi, O., and Mueller, E.~T.
\newblock 2019.
\newblock Dlgnet: A transformer-based model for dialogue response generation.

\bibitem[\protect\citeauthoryear{Peng \bgroup et al\mbox.\egroup
  }{2020a}]{Bao20soloist}
Peng, B.; Li, C.; Li, J.; Shayandeh, S.; Liden, L.; and Gao, J.
\newblock 2020a.
\newblock Soloist: Few-shot task-oriented dialog with a single pre-trained
  auto-regressive model.

\bibitem[\protect\citeauthoryear{Peng \bgroup et al\mbox.\egroup
  }{2020b}]{bao20scgpt}
Peng, B.; Zhu, C.; Li, C.; Li, X.; Li, J.; Zeng, M.; and Gao, J.
\newblock 2020b.
\newblock Few-shot natural language generation for task-oriented dialog.

\bibitem[\protect\citeauthoryear{Peng \bgroup et al\mbox.\egroup
  }{2020c}]{peng2020few}
Peng, B.; Zhu, C.; Li, C.; Li, X.; Li, J.; Zeng, M.; and Gao, J.
\newblock 2020c.
\newblock Few-shot natural language generation for task-oriented dialog.
\newblock {\em arXiv preprint arXiv:2002.12328}.

\bibitem[\protect\citeauthoryear{Peters \bgroup et al\mbox.\egroup
  }{2018}]{peters18}
Peters, M.; Neumann, M.; Iyyer, M.; Gardner, M.; Clark, C.; Lee, K.; and
  Zettlemoyer, L.
\newblock 2018.
\newblock Deep contextualized word representations.
\newblock In {\em Proceedings of the 2018 Conference of the North {A}merican
  Chapter of the Association for Computational Linguistics: Human Language
  Technologies, Volume 1 (Long Papers)},  2227--2237.
\newblock New Orleans, Louisiana: Association for Computational Linguistics.

\bibitem[\protect\citeauthoryear{Radford and Salimans}{2018}]{GPT2pre}
Radford, A., and Salimans, T.
\newblock 2018.
\newblock Improving language understanding by generative pre-training.

\bibitem[\protect\citeauthoryear{Radford \bgroup et al\mbox.\egroup
  }{2018}]{radford2018improving}
Radford, A.; Narasimhan, K.; Salimans, T.; and Sutskever, I.
\newblock 2018.
\newblock Improving language understanding by generative pre-training.

\bibitem[\protect\citeauthoryear{Raffel \bgroup et al\mbox.\egroup
  }{2019}]{raffel2019exploring}
Raffel, C.; Shazeer, N.; Roberts, A.; Lee, K.; Narang, S.; Matena, M.; Zhou,
  Y.; Li, W.; and Liu, P.~J.
\newblock 2019.
\newblock Exploring the limits of transfer learning with a unified text-to-text
  transformer.

\bibitem[\protect\citeauthoryear{Serban \bgroup et al\mbox.\egroup
  }{2016}]{Serban2016BuildingED}
Serban, I.; Sordoni, A.; Bengio, Y.; Courville, A.~C.; and Pineau, J.
\newblock 2016.
\newblock Building end-to-end dialogue systems using generative hierarchical
  neural network models.
\newblock In {\em AAAI}.

\bibitem[\protect\citeauthoryear{Serban \bgroup et al\mbox.\egroup
  }{2017}]{serban17}
Serban, I.~V.; Sordoni, A.; Lowe, R.; Charlin, L.; Pineau, J.; Courville, A.;
  and Bengio, Y.
\newblock 2017.
\newblock A hierarchical latent variable encoder-decoder model for generating
  dialogues.
\newblock In {\em Proceedings of the Thirty-First AAAI Conference on Artificial
  Intelligence}, AAAI'17,  3295–3301.
\newblock AAAI Press.

\bibitem[\protect\citeauthoryear{Shrestha and McKeown}{2004}]{Mckewon2004}
Shrestha, L., and McKeown, K.
\newblock 2004.
\newblock Detection of question-answer pairs in email conversations.
\newblock In {\em Proceedings of the 20th International Conference on
  Computational Linguistics}, COLING '04,  889–es.
\newblock USA: Association for Computational Linguistics.

\bibitem[\protect\citeauthoryear{Sordoni \bgroup et al\mbox.\egroup
  }{2015}]{aless2015hierarchical}
Sordoni, A.; Bengio, Y.; Vahabi, H.; Lioma, C.; Simonsen, J.~G.; and Nie, J.-Y.
\newblock 2015.
\newblock A hierarchical recurrent encoder-decoder for generative context-aware
  query suggestion.

\bibitem[\protect\citeauthoryear{Wen \bgroup et al\mbox.\egroup
  }{2017}]{wenN2N17}
Wen, T.-H.; Vandyke, D.; Mrk\v{s}i\'{c}, N.; Gasic, M.; Rojas~Barahona, L.~M.;
  Su, P.-H.; Ultes, S.; and Young, S.
\newblock 2017.
\newblock A network-based end-to-end trainable task-oriented dialogue system.
\newblock In {\em EACL},  438--449.
\newblock Valencia, Spain: Association for Computational Linguistics.

\bibitem[\protect\citeauthoryear{Wolf \bgroup et al\mbox.\egroup
  }{2019a}]{wolf2019huggingfaces}
Wolf, T.; Debut, L.; Sanh, V.; Chaumond, J.; Delangue, C.; Moi, A.; Cistac, P.;
  Rault, T.; Louf, R.; Funtowicz, M.; Davison, J.; Shleifer, S.; von Platen,
  P.; Ma, C.; Jernite, Y.; Plu, J.; Xu, C.; Scao, T.~L.; Gugger, S.; Drame, M.;
  Lhoest, Q.; and Rush, A.~M.
\newblock 2019a.
\newblock Huggingface's transformers: State-of-the-art natural language
  processing.

\bibitem[\protect\citeauthoryear{Wolf \bgroup et al\mbox.\egroup
  }{2019b}]{Wolf2019TransferTransfo}
Wolf, T.; Sanh, V.; Chaumond, J.; and Delangue, C.
\newblock 2019b.
\newblock Transfertransfo: {A} transfer learning approach for neural network
  based conversational agents.
\newblock {\em CoRR} abs/1901.08149.

\bibitem[\protect\citeauthoryear{Wu \bgroup et al\mbox.\egroup
  }{2020a}]{wu2020todbert}
Wu, C.-S.; Hoi, S.; Socher, R.; and Xiong, C.
\newblock 2020a.
\newblock Tod-bert: Pre-trained natural language understanding for
  task-oriented dialogues.

\bibitem[\protect\citeauthoryear{Wu \bgroup et al\mbox.\egroup
  }{2020b}]{wu2020controllable}
Wu, Z.; Galley, M.; Brockett, C.; Zhang, Y.; Gao, X.; Quirk, C.;
  Koncel-Kedziorski, R.; Gao, J.; Hajishirzi, H.; Ostendorf, M.; and Dolan, B.
\newblock 2020b.
\newblock A controllable model of grounded response generation.

\bibitem[\protect\citeauthoryear{Yang \bgroup et al\mbox.\egroup
  }{2019a}]{yang19}
Yang, Z.; Dai, Z.; Yang, Y.; Carbonell, J.; Salakhutdinov, R.; and Le, Q.~V.
\newblock 2019a.
\newblock Xlnet: Generalized autoregressive pretraining for language
  understanding.
\newblock In {\em arXiv preprint arXiv:1906.08237}.

\bibitem[\protect\citeauthoryear{Yang \bgroup et al\mbox.\egroup
  }{2019b}]{Yang2019xlnet}
Yang, Z.; Dai, Z.; Yang, Y.; Carbonell, J.~G.; Salakhutdinov, R.; and Le, Q.~V.
\newblock 2019b.
\newblock Xlnet: Generalized autoregressive pretraining for language
  understanding.
\newblock {\em CoRR} abs/1906.08237.

\bibitem[\protect\citeauthoryear{Zellers \bgroup et al\mbox.\egroup
  }{2019}]{zellers19}
Zellers, R.; Holtzman, A.; Rashkin, H.; Bisk, Y.; Farhadi, A.; Roesner, F.; and
  Choi, Y.
\newblock 2019.
\newblock Defending against neural fake news.
\newblock {\em CoRR} abs/1905.12616.

\bibitem[\protect\citeauthoryear{Zhang \bgroup et al\mbox.\egroup
  }{2019a}]{Zhang2019ReCoSa}
Zhang, H.; Lan, Y.; Pang, L.; Guo, J.; and Cheng, X.
\newblock 2019a.
\newblock Recosa: Detecting the relevant contexts with self-attention for
  multi-turn dialogue generation.
\newblock {\em CoRR} abs/1907.05339.

\bibitem[\protect\citeauthoryear{Zhang \bgroup et al\mbox.\egroup
  }{2019b}]{zhang20dialogpt}
Zhang, Y.; Sun, S.; Galley, M.; Chen, Y.-C.; Brockett, C.; Gao, X.; Gao, J.;
  Liu, J.; and Dolan, B.
\newblock 2019b.
\newblock Dialogpt: Large-scale generative pre-training for conversational
  response generation.

\end{thebibliography}

\end{document}